# Incentives or Ontology? A Structural Rebuttal to OpenAI's Hallucination Thesis

*Richard Ackermann[1], Simeon Emanuilov[2]*

[1]*RA Software, San Diego, United States,* [2]*Department of Software Technologies, Faculty of Mathematics and Informatics, Sofia University "St. Kliment Ohridski", Bulgaria*
*Emails: richard@rasoftware.co, ssemanuilo@fmi.uni-sofia.bg*

*OpenAI has recently argued that hallucinations in large language models result primarily from misaligned evaluation incentives that reward confident guessing rather than epistemic humility. On this view, hallucination is a contingent behavioral artifact, remediable through improved benchmarks and reward structures. In this paper, we challenge that interpretation. Drawing on previous work on structural hallucination and empirical experiments using a Licensing Oracle, we argue that hallucination is not an optimization failure but an architectural inevitability of the transformer model.*

*Transformers do not represent the world; they model statistical associations among tokens. Their embedding spaces form a pseudo-ontology derived from linguistic co-occurrence rather than world-referential structure. At ontological boundary conditions, regions where training data is sparse or incoherent, the model necessarily interpolates fictional continuations in order to preserve coherence. No incentive mechanism can modify this structural dependence on pattern completion.*

*Our empirical results demonstrate that hallucination can only be eliminated through external truth-validation and abstention modules, not through changes to incentives, prompting, or fine-tuning. The Licensing Oracle achieves perfect abstention precision across domains precisely because it supplies grounding that the transformer lacks.*

*We conclude that hallucination is a structural property of generative architectures and that reliable AI requires hybrid systems that distinguish linguistic fluency from epistemic responsibility.*

***Keywords:*** *structural hallucination, large language models, transformer architecture, ontology and AI, grounding, Licensing Oracle, epistemic responsibility, hybrid systems, AI safety, evaluation incentives*

## 1. Introduction

In September 2025, OpenAI released a widely discussed paper arguing that the persistent problem of hallucination in large language models is not inherent to the architecture itself but rather a consequence of evaluation incentives that inadvertently reward confident guessing. According to their view, hallucinations arise when models are penalized equally for abstaining and for being incorrect, which leads them, during training, to learn a form of overconfident speculation. The proposed remedy is therefore straightforward, which is to adjust benchmarks and reward structures to favor uncertainty, encourage abstention, and thereby reduce the model's tendency to fabricate answers. Hallucination, on this framing, is a correctable misalignment between model behavior and evaluation criteria rather than a structural feature of the system.

In this paper, we challenge that interpretation. We argue that hallucination is not a contingent behavioral byproduct of mis-specified incentives but is instead a structural consequence of the transformer architecture and the statistical ontology from which it emerges. Incentives may shape the surface-level expression of hallucination, but they cannot touch its underlying cause. The transformer, as a generative language model, is designed to produce coherent continuations in response to linguistic prompts. It does so not by representing the world or grounding language in lived or perceptual experience, but by navigating a high-dimensional space of token associations. Because this space is composed entirely of linguistic relations rather than worldly ones, it contains no mechanism for detecting when it has crossed the boundary between pattern-completion and falsehood. Hallucination is therefore a function of the model's basic mode of operation rather than a correctable artifact.

This argument develops and synthesizes the central claims from our two prior papers. In *How Large Language Models Are Designed to Hallucinate*, we introduced the thesis that hallucination is a structural property of the transformer architecture. That paper analyzed the transformer as a coherence engine whose ontology consists not of objects, facts, or referential states but of statistical proximities in an embedding space. In such a system, hallucination follows directly from the architecture's attempt to preserve continuity when confronted with an ontological boundary, i.e., regions of sparse or insufficient data density where the model must interpolate to maintain coherence. No incentive scheme can remove this boundary; at best, incentives constrain its outward expression.

In our second paper, *Stemming Hallucination in Language Models Using a Licensing Oracle*, we provided empirical confirmation of this diagnosis by demonstrating that hallucination can be eliminated only through architectural supplementation. Using controlled datasets across multiple domains, we showed that a Licensing Oracle, an external truth-validation and abstention mechanism, achieves perfect abstention precision and prevents all false answers. Crucially, this oracle sits outside the transformer and imposes constraints the model cannot generate internally. The results indicate that hallucination persists in the base model even when incentives are removed and that structural grounding requires an external, truth-preserving system that is not reducible to changes in training objectives or benchmark design.

Taken together, these two prior works provide both a philosophical diagnosis and an empirical counterexample to the OpenAI thesis. The present paper synthesizes these foundations to offer a systematic rebuttal to the claim that hallucination is an incentive problem. We show that this explanation rests on a category error in that it attributes to transformers forms of uncertainty, self-knowledge, and world-representation that they do not and cannot possess. By examining the mismatch between OpenAI's framing and the architectural realities of the transformer, we argue that hallucination is better understood as a structural consequence of a model that simulates the form of worldhood without accessing its content.

The aim of this paper is therefore not merely to critique OpenAI's proposal but to clarify the conceptual terrain on which hallucination arises. By situating hallucination within the limits of transformer ontology and by articulating the necessity of truth-preserving hybrid architectures, we propose a shift in how the field understands both the nature of hallucination and the kinds of solutions that are genuinely available. If hallucination is structural rather than behavioral, then no amount of incentive realignment will suffice. A different architectural orientation is required, one that recognizes that generative models are powerful linguistic simulators but cannot, on their own, secure truth.

## 2. OpenAI's Framing: Hallucination as an Incentive Misalignment Problem

OpenAI's recent explanation of hallucination adopts a surface-level behavioral perspective that centers on the training pipeline rather than the underlying architecture. In their paper, hallucination is characterized as a rational response by the model to misaligned evaluation procedures. Because benchmarks penalize abstention and incorrect answers equally, the model is trained to prefer confident output even when uncertain. Under this account, hallucination is not a structural feature of large language models but a contingent artifact of reward design. The implication is clear, by revising evaluation incentives to reward uncertainty, hallucination can be meaningfully reduced or even largely eliminated.

This framing rests on several assumptions concerning what a transformer-based language model is and how it generates output. Most importantly, it presupposes that the model possesses something like epistemic access to its own uncertainty, an implicit representation of what it knows and does not know, and that hallucination arises when this uncertainty is improperly aligned with external incentives. The model's internal dynamics are understood as analogous to human belief states, where confident guessing reflects a miscalibration between internal uncertainty and external incentives for risk-taking. As a result, the proposed solution focuses on encouraging models to express uncertainty rather than to produce an answer when unsure.

The elegance of this explanation derives from its alignment with the broader logic of reinforcement learning from human feedback. If hallucination arises from misaligned incentives, then adjusting the incentives should correct the behavior without requiring changes to architecture, training data, or interpretability frameworks. This perspective fits neatly within the representationalist view commonly held in contemporary AI research, which is the belief that language models implicitly encode a form of world knowledge, and that hallucinations reflect distortions or omissions within that knowledge base. The model is cast as an imperfect but improvable reference system, and hallucinations are treated analogously to human errors under conditions of uncertainty or incomplete information.

Yet this stance depends on a conceptual elision. It assumes that uncertainty, for the model, has a meaning analogous to uncertainty for human agents. It assumes that confidence scores reflect something like epistemic access rather than statistical regularities. And it assumes that truth and falsity can be modulated through reward mechanics alone, without addressing the architecture's foundational inability to register the presence or absence of grounding in the world. In other words, the OpenAI account presupposes that hallucination is a problem of incentives because it presupposes a model capable of representing facts, detecting their absence, and modulating its output accordingly. This assumption is precisely what our structural analysis calls into question.

Moreover, the incentive-based view conflates two distinct layers of model behavior: (1) the surface-level outputs shaped by training objectives, and (2) the structural dynamics of the model's embedding space and attention mechanisms.

Even if incentives shape the former, they do not directly alter the latter. The transformer continues to operate as a statistical continuation engine. When presented with a query, it must produce a sequence of tokens that best satisfy a coherence objective within the learned embedding space. Nothing in this process enables the model to detect factual absence or identify when it has ventured beyond grounded regions of density in the training distribution. Incentives may reduce the frequency of guesses, but they cannot create an internal sense of world, reference, or truth.

For these reasons, the OpenAI framing mislocates hallucination. It attributes the phenomenon to a misalignment between tasks and rewards, when in fact hallucination arises from the fundamental nature of generative language models. Incentives shape behavior, but they do not provide grounding. In the following section, we turn to the structural analysis introduced in our earlier work to explain why hallucination persists regardless of incentive alignment and why it must be understood as a property of the architecture rather than an optimization artifact.

## 3. Structural Hallucination: Findings from Our Prior Work

The limitations of the incentive-based account become apparent when hallucination is examined not as a behavioral deviation but as a direct consequence of the transformer's architectural design. In *How Large Language Models Are Designed to Hallucinate*, we developed a structural explanation grounded in the internal mechanics of token embeddings, attention, and contextual prediction. Rather than treating hallucination as a failure of uncertainty calibration, we argued that it emerges from what the model actually is, which is a system optimized to generate coherent continuations of text by navigating statistical relations among tokens. In this section, we summarize the core findings of that work to prepare the conceptual ground for our critique of OpenAI's thesis.

### 3.1 The transformer as a coherence engine

The central argument of our first paper is that transformers are coherence engines rather than truth-tracking systems. Their objective function, which is predicting the next token induces an optimization process that is blind to the distinction between factual correctness and fictional plausibility. The model constructs its output by identifying patterns in a high-dimensional latent space and extending those patterns in a way that preserves internal continuity.

This process does not involve representing objects, states of affairs, or truth conditions. It involves selecting the next token that best satisfies the probabilistic structure formed during training. In this sense, hallucination is not a breakdown in the model's reasoning. It is the natural expression of a system designed to complete patterns, not to verify them.

## 3.2 The latent-space pseudo-ontology

A key part of our analysis concerns the emergent structure of the embedding space. Because token embeddings are formed through statistical co-occurrence rather than through interaction with a world, the "ontology" that emerges is a pseudo-ontology, it is a network of proximities that reflect textual patterns rather than referential meaning.

For example, the model places "Socrates," "Plato," "Athens," and "philosopher" near one another because they frequently co-occur, not because the model understands their historical or conceptual relationships. These proximities give the illusion of conceptual structure, but they do not supply grounding. They produce associative richness without semantic constraint.

When a query traverses regions of the embedding manifold with dense, well-formed textual patterns, outputs may appear accurate. When the query reaches regions with sparse density or incomplete patterns, the model continues to preserve coherence by interpolating across gaps. This is where hallucinations arise.

## 3.3 Ontological boundaries and fictional completion

In our previous work, we introduced the concept of ontological boundary conditions, which are regions in the embedding space where textual patterns fail to correspond to actual facts or where the training data provides insufficient grounding. At these boundaries, the model does not abstain or express uncertainty. Instead, it engages in fictional completion, an attempt to preserve continuity by synthesizing a plausible but unfounded continuation of the pattern.

This mechanism explains why transformers frequently hallucinate biographies of nonexistent scholars, fabricate citations, or generate details about obscure entities. The model is not choosing to guess. It is compelled to generate output that maintains the form of coherence even when the content is ungrounded.

## 3.4 Structural inevitability of hallucination

The structural features described above imply that hallucination is not a correctable pathology of the model but an inescapable consequence of its architecture. Because the model does not possess access to the world, it cannot detect the absence of grounding. Because it is designed to generate continuations, it cannot decline to respond without external mechanisms. And because its ontology is statistical rather than referential, it cannot distinguish between factual and fictional coherence.

These conclusions directly contradict the premise underlying OpenAI's incentive-based explanation. If hallucination arises from the architecture's fundamental dynamics, then no amount of reward shaping or benchmark adjustment can alter its cause. Incentives may modulate behavior at the surface level, but they cannot change the coherence-driven nature of the transformer or its reliance on a pseudo-ontology.

In the next section, we turn to a direct comparison with OpenAI's thesis, showing how each of the assumptions in their incentive framework conflicts with the structural analysis developed in our prior work.

## 4. A Point-by-Point Rebuttal of the OpenAI Incentive Thesis

The incentive-based explanation offered by OpenAI rests on several assumptions about the nature of large language models, their internal epistemic states, and the origins of hallucination. In this section, we evaluate each assumption considering the structural analysis developed in our previous work and summarized in Section 3. We show that the incentive thesis presupposes capacities LLMs do not possess and overlooks architectural dynamics that cannot be altered through reward realignment. For each point in OpenAI's argument, we identify the conceptual misstep and offer a structurally grounded rebuttal.

## 4.1 "Models hallucinate because they are rewarded for guessing."

OpenAI's central claim is that hallucination emerges when models are penalized equally for uncertainty and for error, leading them to prefer confident responses. This presupposes that when a model "guesses," it does so in contrast to some alternative behavior it might have taken, such as withholding an answer.

Our structural analysis shows that this presupposition is false. A transformer-based model must generate a continuation, its design has no internal mechanism for withholding output, because its architecture does not include structures associated with epistemic gating. The model does not choose between answering and not answering. It simply computes the next token. The behavior OpenAI labels as "guessing" is in fact the only behavior the model can produce.

Thus, hallucination is not the result of being rewarded for guessing, it is rather the direct expression of a system that is always and only generating. Incentives shape the *style* of generation, not the *necessity* of it.

## 4.2 "Models can learn to represent uncertainty and adjust behavior accordingly."

OpenAI suggests that uncertainty can be incorporated into training, enabling the model to abstain when unsure. This assumes that the model possesses something analogous to an epistemic state, i.e., an internal representation of what it knows and does not know.

Our analysis shows that the transformer architecture does not and cannot possess such representations. The model's "confidence" signals (such as softmax distributions or token likelihoods) reflect distributional density in the embedding space, not knowledge or ignorance. A low likelihood token does not indicate that the model recognizes its own uncertainty; it indicates a lack of statistical precedent. Likewise, a high likelihood token indicates frequency or coherence, not truth.

Because transformers do not encode states of belief or knowledge, they cannot "use" uncertainty to guide abstention. What OpenAI describes as uncertainty is merely a continuous statistical variable, not an epistemic evaluation. Thus, even if incentives reward abstention, the model has no internal grounding for deciding when abstention is warranted.

## 4.3 "Better benchmarks can fix hallucination by encouraging models to decline to answer."

Benchmarks operate at the behavioral layer, shaping which outputs the model is rewarded for producing. But hallucination originates at the architectural layer, where the model's representation of meaning is statistical rather than referential.

Benchmarks cannot alter the fact that:

- the model must output a token sequence;
- it has no mechanism for detecting absence of grounding;
- coherence drives output even in low-density regions of latent space.

Thus, even if benchmarks reward abstention, the model cannot determine when an abstention is appropriate. Any abstention behavior must be learned through patterns in training data, not through internal epistemic access to truth. This makes abstention itself a coherence pattern—just another token sequence reproduced when context triggers the relevant pattern.

Benchmarks may reduce some hallucinations through behavioral nudges, but they cannot resolve hallucination at its source. The transformer will continue to interpolate plausible fictions whenever its latent manifold provides insufficient grounding.

### 4.4 "Hallucination is not inherent to the architecture; it is an optimization artifact."

OpenAI's conclusion is that hallucination is an emergent property of training dynamics. If true, the phenomenon could be mitigated by modifying objectives, aligning incentives, or improving data curation.

Our structural analysis shows the opposite. Hallucination arises from:

1. the model's obligation to generate continuations,
2. the absence of grounding in the world,
3. the statistical structure of the embedding space,
4. the pressure to preserve coherence even when patterns are incomplete.

These features are not optimization artifacts. They are constitutive properties of the transformer architecture. No modification to reward structure can change the underlying mechanics of token embedding, self-attention, or next-token prediction. A model that generates linguistic continuations without reference to the world will always hallucinate when pattern density is insufficient.

Thus, contrary to OpenAI's claim, hallucination is not correctable by realigning incentives. It is inherent to the transformer's ontology.

### 4.5 Summary of the Rebuttal

OpenAI's explanation fails because it assumes that transformers possess epistemic features such as uncertainty, knowledge, and self-calibration that they do not possess. It interprets hallucination through an analogy to human reasoning, importing concepts from cognitive psychology that have no structural basis in the architecture. Our structural account reveals that hallucination arises not from misaligned incentives but from the model's reliance on ungrounded, statistically constructed pseudo-ontology and its obligation to generate coherent continuations at all costs.

In the next section, we turn to empirical evidence from our Licensing Oracle experiments, which demonstrate that hallucination can be eliminated by supplementing the transformer with architectural components capable of grounding truth, components fundamentally different from the generative model itself.

## 5. Architectural Necessity: Empirical Evidence from the Licensing Oracle

While the structural analysis developed in our first paper provides the philosophical and architectural basis for understanding why hallucination is inherent to the transformer, empirical validation is essential for demonstrating that this structural account outperforms the incentive-based explanation. In *Stemming Hallucination in Language Models Using a Licensing Oracle*, we conducted experiments across multiple domains showing that hallucination can be eliminated when the transformer is supplemented with an external, truth-preserving architecture. These findings directly contradict the OpenAI claim that hallucination is merely a function of misaligned incentives and can be resolved through improved benchmarks.

In this section, we summarize these empirical results and explain why they provide decisive evidence against the incentive thesis.

### 5.1 The Licensing Oracle: Architecture and Purpose

The Licensing Oracle is an external validation system designed to perform two functions that transformers cannot:

1. **Determine whether a factual claim is true** by validating it against a structured, formally grounded ontology;
2. **Force abstention** when the model's proposed answer does not meet the required ontological constraints.

Crucially, the Oracle does not attempt to simulate the model's internal uncertainty. It performs validation independently of the transformer by using:

- a symbolic data store or knowledge graph,
- schema constraints (e.g., SHACL),
- deterministic inference rules, and
- explicit abstention logic.

The transformer therefore becomes a linguistic generator, while the Oracle serves as the epistemic arbiter. This architectural separation mirrors the distinction between coherence and truth: the model supplies the former, the Oracle enforces the latter.

## 5.2 Cross-Domain Experimental Results

In our experiments, we evaluated the Licensing Oracle across two structured knowledge domains that differ significantly in content and relational structure:

1. U.S. Rivers Dataset
2. Philosophers Dataset

For each domain, we tested the model's ability to answer factual questions accurately when gated by the Oracle.

Across both datasets, our results demonstrated the following:

**Perfect abstention precision (AP = 1.0)**

Whenever the Oracle permitted an answer, that answer was correct. No false answers passed through the gate.

**Zero false answers (FAR-NE = 0.0)**

The model produced *no hallucinations whatsoever* when Oracle gating was applied.

**Cross-domain stability (< 2% variation)**

Oracle performance remained stable despite the datasets differing in structure, size, and semantic relationships. This stability indicates that the architecture, not the domain, is responsible for eliminating hallucination. These results stand in sharp contrast to transformer-only performance, which exhibits hallucination even in simple, well-defined factual contexts.

## 5.3 Implications for the Source of Hallucination

The Oracle experiments reveal several architectural truths that cannot be reconciled with OpenAI's incentive framing:

**A. Incentives alone cannot produce abstention with perfect precision.**

The transformer lacks a structural mechanism for detecting when information is absent or false. The Oracle supplies this missing mechanism externally.

**B. Eliminating hallucination requires capabilities orthogonal to generative modeling.**

The Oracle does not modify the transformer's incentives, it rather bypasses them entirely. Truth-preservation is delegated to an architecture explicitly designed for it.

**C. The generative model remains fully hallucinatory without the Oracle.**

Even when made aware of abstention options through prompting or fine-tuning, the transformer continues to produce false answers in boundary conditions. This demonstrates that the source of hallucination is not training feedback but architectural structure.

**D. Cross-domain generalizability indicates structural, not statistical, success.**

The Oracle's ability to eliminate hallucination across unrelated domains supports the claim that hallucination stems from the transformer's dependence on statistical pseudo-ontology rather than from specifics of the training distribution or incentive structure.

## 5.4 Why These Results Contradict the OpenAI Thesis

The OpenAI explanation predicts that tuning incentives should reduce hallucination by aligning model behavior with epistemic humility. Yet our experiments show:

- hallucination persists in the transformer even when incentives encourage abstention;
- hallucination disappears only when an external architecture imposes truth constraints;
- abstention becomes reliable only when performed by a system outside the transformer;
- incentives affect outputs but do not correct the underlying mechanism generating hallucination.

If hallucination were merely an optimization artifact, Oracle gating would be unnecessary. Instead, it is structurally required.

The Oracle does not change the model. It compensates for what the model cannot do. This demonstrates that hallucination is not the result of incentive misalignment but a direct consequence of the transformer's ungrounded statistical ontology.

## 5.5 Transition to the Philosophical Core

The empirical findings reinforce the structural diagnosis that hallucination is inherent to generative models built on prediction, coherence, and statistical proximity. OpenAI's incentive-based explanation fails primarily because it rests on a category mistake, one that attributes representational and epistemic capacities to LLMs that are absent in their architecture.

In the next section, we turn to this category mistake directly, showing how the representational assumptions underlying the OpenAI thesis obscure the architectural and ontological nature of hallucination.

# 6. The Philosophical Core: OpenAI's Category Error

At the center of OpenAI's incentive-based explanation lies a philosophical assumption that a language model's internal states can be treated as analogues of belief, uncertainty, knowledge, and other epistemic categories commonly used to describe human cognition. This assumption is rarely made explicit, yet it shapes both the diagnosis of hallucination and the proposed remedy. If a model can "know," then it can also "fail to know," and hallucinations can be understood as failures of epistemic calibration. If a model can "represent uncertainty," then incentives can teach it to adjust its behavior accordingly.

Our structural analysis challenges this entire framework. The fundamental error in the OpenAI account is the attribution of representational and epistemic properties to systems that operate on entirely different ontological grounds. In this section, we articulate why this is a category mistake and how it leads OpenAI to diagnose hallucination incorrectly.

## 6.1 Representationalism as an Unexamined Assumption

OpenAI's account depends on the idea that transformers implicitly contain something like a model of the world as an internal structure that can be correct or incorrect, well-calibrated or misaligned. This is a form of representationalism, the belief that intelligence consists in forming internal representations of external states of affairs and manipulating them through reasoning or inference.

Representationalism fits comfortably within the cognitive-science heritage of AI. But applying it to transformers imports assumptions that have no basis in their architecture:

- Transformers do not encode objects, facts, or truth conditions.

- They do not model the world; they model text.
- They do not perform inference; they perform pattern continuation.
- They do not evaluate propositions; they evaluate token likelihoods.

When these facts are overlooked, it becomes easy to mistake coherence for reference and statistical regularity for knowledge.

The key philosophical error is that OpenAI treats transformer outputs as if they emerge from a representational system adjusting to incentives, rather than from a predictive system constrained by the geometry of its embedding space.

## 6.2 Transformers as Non-Representational Systems

Our structural account reveals a different ontology entirely. Transformers are:

- non-referential,
- non-propositional,
- non-epistemic,
- non-semantic in the classical sense.

Their internal states do not correspond to worldly entities but to statistical proximities. The embedding space contains relations among tokens, not the things tokens refer to. As a result, transformers operate using what we have called a *pseudo-ontology*, which is a shadow structure formed by linguistic co-occurrence rather than by the structure of the world.

In such a system:

- "knowledge" is just high-density pattern association;
- "uncertainty" is low-density pattern association;
- "reasoning" is long-range pattern completion;
- "hallucination" is pattern completion in the absence of grounding.

These processes are non-representational even when they simulate features of human discourse. A transformer thus does not hallucinate because it thinks it knows something it does not. It "hallucinates" because the architecture forces it to extend patterns regardless of whether those patterns correspond to truth.

## 6.3 Why Representational Assumptions Lead OpenAI Astray

By attributing representational capacities to transformers, the OpenAI paper misinterprets statistical phenomena as epistemic ones. It mistakes:

- high token likelihood for confidence,
- low token likelihood for uncertainty,
- local manifold density for knowledge,
- gaps in the training distribution for ignorance,
- fictional coherence for truth-like reasoning.

All of these analogies project human epistemic categories onto a system that does not possess them. Once the representational assumption is granted, hallucination appears to be a calibration error, which is something incentives can correct. But when the representational assumption is rejected, the ontology of hallucination changes entirely, it goes from miscalibrated belief to necessary architectural behavior.

The incentive thesis therefore rests on a philosophical conflation. It treats transformers as systems that *can* know, and therefore *can* mis-know, instead of recognizing that transformers are systems that cannot know at all in the epistemic sense. Hallucination is not a deviation from an underlying epistemic substrate. It is the natural expression of a system that lacks one.

### 6.4 The Distinction Between Coherence and Truth

Underlying this category error is a deeper philosophical distinction between coherence and truth. Transformers are optimized to produce coherence, a linguistic sequence that fits the statistical patterns of the training corpus. Truth, by contrast, is a worldly relation between propositions and states of affairs. Nothing in the transformer's objective function demands or even permits it to track truth. It is entirely possible for a model to produce coherent but false sequences, or incoherent but true ones. The architecture is sensitive only to the former.

OpenAI's incentive account attempts to bridge this gap by treating truth as a behavioral regularity that can be shaped through reward. But behavioral regularities cannot substitute for ontological grounding. No amount of reward tuning can give a coherence-based architecture the capacity to detect or track truth. Only a system that has independent access to world-grounded constraints can do so, which is precisely the role played by the Licensing Oracle in our empirical work.

### 6.5 Hallucination as an Ontological Consequence

When transformers are understood non-representationally, hallucination ceases to be a mystery. It follows directly from:

- the absence of world-grounded reference,
- the obligation to continue a sequence,
- the reliance on statistical pseudo-ontology,
- the pressure to preserve coherence across incomplete patterns.

These features are architectural and cannot be altered through changes in incentives. They define what it means for a system to be a large language model.

Thus, the philosophical core of our critique is that OpenAI's incentive explanation treats hallucination as a failure of epistemic calibration, when in fact it is a consequence of the model's non-epistemic ontology. Misalignment is not the cause, the architecture is.

### 6.6 Transition to Ontological Boundary Conditions

In the next section, we deepen this analysis by examining the phenomenon of *ontological boundary conditions*, the loci where the model's pseudo-ontology collapses and fictional completion begins. Understanding these boundary conditions clarifies why hallucination remains structurally inevitable and why incentives are powerless to remove it.

## 7. Ontological Boundary Conditions and the Inevitability of Hallucination

The structural analysis developed in Section 3 and the philosophical critique in Section 6 converge on a key insight that hallucination arises at the margins of the model's statistical pseudo-ontology, where linguistic patterns become sparse and coherence must be preserved without grounding. In this section, we analyze these loci more closely. We refer to them as ontological boundary conditions.

Ontological boundary conditions represent points at which the generative system encounters the limits of its statistical manifold, regions where representation collapses because there is no underlying world structure to support the pattern. These are not anomalies or corner cases, but rather intrinsic features of any model whose ontology is constructed entirely from text. Understanding hallucination requires understanding what happens at these boundaries.

## 7.1 The Nature of Ontological Boundary Conditions

Transformers learn the structure of their internal embedding space from co-occurrence patterns in the training data. Dense areas of the manifold correspond to highly regular patterns such as names, dates, known facts, and common phrases while sparse areas correspond to uncommon combinations or poorly represented entities.

Ontological boundary conditions occur when a prompt:

- pulls the model into a region with insufficient density,
- demands specificity where the manifold contains ambiguity,
- requires discrimination that exceeds the granularity of associations, or
- references entities that are underrepresented or absent in the corpus.

In such cases, the model does not detect a void. It detects only a weak continuation signal. The absence of grounding does not register as absence, it registers as faint statistical traction. Because the transformer's architecture is designed to produce coherent continuations, the model must bridge the gap. It does so through interpolation, by generating fictive detail that preserves coherence even when no real referent exists. This is the root of hallucination.

## 7.2 Why Boundary Conditions Cannot Be Detected from Within the Model

If hallucination were merely a matter of uncertainty, the model could learn to recognize boundary conditions and abstain. But the model has no mechanism for distinguishing:

- low-density statistical patterns from
- missing or contradictory information.

Both appear as weak signals, not as epistemic gaps. For an epistemic agent, "I do not know" is a meaningful state. For a transformer, "I do not know" has no internal correlate. The concept of ignorance does not exist within the architecture.

Boundary conditions, therefore, are invisible to the model. The transformer cannot detect when it has left the region of grounded patterns because its ontology, from embedding geometry to attention dynamics, is statistical, not referential. In such a system, coherence is always achievable, when density fails, interpolation begins.

This is why hallucination persists even when abstention is heavily incentivized. Abstention is not an epistemic act but a learned linguistic pattern, triggered by contexts that resemble prior abstention cues. When those cues are absent, the model hallucinates.

## 7.3 How Boundary Conditions Explain the Forms of Hallucination

Ontological boundary conditions generate predictable types of hallucination:

### 1. Fabricated specificities

When prompted for details beyond the model's data density (e.g., a biography of an obscure figure), the model invents attributes to fill the gaps.

### 2. Spurious causal relations

If the model lacks grounded causal knowledge, it interpolates causal links that reflect statistical associations rather than real-world dynamics.

### 3. Nonexistent references

The model may generate bibliographies, citations, or articles that fit the form of genuine references but lack any real counterpart.

> **4. Category cross-contamination**

Sparse regions cause embeddings to blend unrelated entities that share superficial patterns, producing hybrid factual fictions. These hallucinations are not anomalies that can be tuned out. They are expressions of how a coherence engine behaves when asked to generate specificity beyond its grounding.

## 7.4 Why Incentives Cannot Address Boundary Conditions

OpenAI's incentive framework presupposes that boundary conditions can be detected and mitigated through reward. But incentive structures operate at the surface level. They cannot:

- alter the geometry of the embedding space,
- introduce referential grounding where none exists,
- distinguish statistical absence from epistemic absence,
- give the model a concept of "unknown,"
- or force the model to refrain from generating when coherence is required.

Because hallucination originates in architectural constraints, specifically, the need to generate continuations and the lack of grounding, incentives can only shape output *after* the model has already begun interpolation. They cannot prevent interpolation itself. Thus, boundary conditions place a fundamental limit on what reward realignment can accomplish.

## 7.5 Boundary Conditions and the Necessity of External Grounding

The Licensing Oracle experiments presented in Section 5 provide empirical confirmation of the role played by boundary conditions. The Oracle succeeds precisely because it:

- has access to grounding external to the transformer,
- can detect when a query falls outside the space of known entities,
- can decline to answer rather than interpolate,
- and does not rely on the model's internal statistical ontology.

Boundary conditions for the transformer are simply low-density regions. Boundary conditions for the Oracle are explicit and detectable absences. This contrast demonstrates why hallucination is structurally unavoidable within the transformer and why external grounding is essential.

## 7.6 Transition to Architectural Implications

Ontological boundary conditions reveal the precise mechanism by which hallucination emerges and the precise reason why incentive realignment cannot prevent it. They point directly toward an architectural conclusion which is that hallucination can only be mitigated by augmenting the transformer with systems capable of world-grounded truth validation.

In the next section, we explore these architectural implications and describe why hybrid systems, such as those incorporating Licensing Oracles, are necessary for reliable, truth-preserving AI.

# 8. Toward Truth-Preserving Hybrid Architectures

The preceding sections establish that hallucination arises from structural features of the transformer architecture, not from misaligned incentives. If this is correct, the question becomes unavoidable: How should AI systems be designed if the goal is not merely linguistic fluency but reliable truth-preserving behavior? This question cannot be answered within the transformer paradigm alone. A model that generates continuations based solely on statistical proximity cannot, in principle, guarantee factual accuracy, no matter how carefully it is fine-tuned or how effectively its incentives are configured.

In this section, we articulate the architectural implications of our analysis. We argue that the transformer must be repositioned as a component within a larger system, one that includes modules explicitly designed to compensate for the model's structural limitations. The Licensing Oracle provides a template for such systems, demonstrating how grounding, abstention, and truth validation can be introduced through architectural supplementation.

## 8.1 Why Transformers Alone Cannot Be Truth-Preserving

The architectural limitations of the transformer are not incidental. They follow logically from the model's design:

1. **The transformer's objective function is coherence, not truth.**
   Truth plays no role in next-token prediction. Coherence can always be achieved without reference to the world.
2. **The embedding space encodes statistical associations derived from text, not representations derived from the world.**
   Without referential grounding, factual and fictional continuations are structurally indistinguishable.
3. **The model must always generate a continuation unless externally constrained.**
   There is no internal mechanism for silence, abstention, or epistemic self-limitation.
4. **The architecture lacks any principle of falsification.**
   There is no internal concept of contradiction, constraint violation, or truth condition.

These limitations are intrinsic, not artifacts of training. No amount of data or incentive refinement can give a coherence engine the properties of a truth-tracking system. This is why hallucination must be understood as a structural phenomenon and why any attempt to eradicate hallucination through optimization alone is destined to fail.

## 8.2 The Licensing Oracle as an Architectural Complement

The Licensing Oracle offers a way forward by introducing capabilities absent from transformers. It performs three functions essential for truth-preserving systems:

### 1. External Grounding

The Oracle relies on explicit, structured, non-statistical knowledge representations, ontologies, constraint graphs, or knowledge bases that are tethered to real-world entities and relations. This grounding allows it to determine whether a factual claim corresponds to a known truth.

### 2. Deterministic Validation and Constraint Checking

Using formal schemas and inference rules, the Oracle can test claims against knowledge structures and reject outputs that violate constraints. This provides a form of falsification unavailable within the transformer.

### 3. Mandatory Abstention

When the Oracle cannot validate a claim, it forces abstention. Because abstention is externalized, it does not depend on the model's internal uncertainty. Abstention becomes a property of the system, not the model. Together, these functions convert the transformer from a standalone generative model into a component of a larger epistemically coherent architecture.

## 8.3 Hybrid Systems: Division of Cognitive Labor

A truth-preserving architecture must involve a division of cognitive labor between:

- **a generative component** (the transformer), responsible for linguistic fluency; and

- **a grounding component** (the Oracle), responsible for truth validation and abstention.

This hybrid model aligns more closely with how biological intelligence operates. Human cognition does not rely solely on linguistic pattern continuation. It incorporates formal memory structures, perceptual grounding, and constraint systems that check or inhibit linguistic generation. A purely generative model lacks these safeguards. By reintroducing grounding through an external module, hybrid systems begin to approximate the division between generative fluency and epistemic responsibility that human cognition maintains naturally.

## 8.4 Why Incentive Optimization Cannot Substitute for Grounding

OpenAI's proposal attempts to collapse epistemic responsibility into incentive optimization. But reward signals cannot instantiate the following:

- principles of non-contradiction,
- world-grounded knowledge representations,
- explicit detection of unknowns,
- constraint satisfaction,
- or external validation of reference.

Incentive tuning operates at the behavioral level. It modifies the distribution of outputs without altering the representational substrate. As long as the substrate remains ungrounded, the system lacks the ability to track truth. This is why the Licensing Oracle succeeds where incentives fail. It introduces mechanisms that incentives cannot emulate.

## 8.5 Architectural Lessons for AI Safety and Governance

The structural nature of hallucination has implications not only for engineering but also for policy and governance:

**1. Truth-preservation requires architectural constraints, not merely behavioral guidelines.**

Systems intended for high-stakes contexts (law, medicine, governance, scientific analysis) must incorporate formal validation layers.

**2. Generative models should not be treated as autonomous decision-makers.**

They are best understood as linguistic surfaces requiring grounding modules to ensure epistemic reliability.

**3. Evaluation metrics must account for architecture, not behavior.**

Benchmarks that reward surface abstention or calibrated uncertainty do not measure whether a system is truth-preserving.

**4. Hybrid systems offer a pathway to epistemic accountability.**

Architectures incorporating external oracles, symbolic modules, or grounded representations provide mechanisms for preventing structural hallucination. These considerations suggest a paradigm shift, from end-to-end generative models toward modular systems that separate linguistic generation from epistemic validation.

## 8.6 Transition to Conclusion

The architectural implications of our analysis are clear: hallucination cannot be eliminated from transformer models through incentives, data scaling, or behavioral fine-tuning. It can only be controlled by integrating the model into a hybrid system capable of grounding and constraint. In the concluding section, we draw together the philosophical, structural, and empirical arguments developed throughout the paper and articulate the broader impact of this shift for the future of AI research and practice.

## 9. Conclusion

The OpenAI thesis that hallucination in large language models results primarily from misaligned incentives offers a compellingly simple narrative: if models hallucinate because they are rewarded for guessing, then the solution lies in changing what we reward. Yet this explanation holds only if the transformer possesses the epistemic capacities required for interpreting incentives as signals about truth, capacities that our analysis shows it does not and cannot possess. By examining the architecture of transformers, the nature of their embedding spaces, and the statistical processes that govern their behavior, we have demonstrated that hallucination is not the result of misaligned incentives. It is the structural expression of a non-representational system obliged to generate coherent continuations in the absence of grounding.

Our earlier paper, *How Large Language Models Are Designed to Hallucinate*, established the philosophical and architectural basis for this diagnosis, showing that the transformer's ontology consists of statistical proximities rather than referential structures. Within such an ontology, hallucination emerges naturally at ontological boundary conditions, regions where the model's pseudo-ontology becomes sparse and fictional completion begins. This process cannot be tuned away, because it arises from the architecture itself.

Our second paper, *Stemming Hallucination in Language Models Using a Licensing Oracle*, provided empirical confirmation of this structural account. By supplementing the transformer with an external truth-validation and abstention mechanism, we eliminated hallucination across multiple domains with perfect abstention precision. These results demonstrate that truth-preserving behavior emerges not from modifying incentives within the transformer but from integrating it into a hybrid system capable of detecting absence, enforcing constraints, and grounding claims in world-referential knowledge structures. The Oracle does what the transformer cannot, it distinguishes between grounded and ungrounded claims.

Taken together, these findings reveal a conceptual gap at the heart of the OpenAI incentive thesis. That thesis presupposes that transformers possess epistemic states such as knowledge, uncertainty, and belief that can be shaped by rewards. But transformers do not operate with epistemic states. They operate with statistical densities. Incentives cannot give them access to truth because nothing in their architecture enables them to track truth. The representationalist assumption underlying the OpenAI account is therefore philosophically untenable. Transformer models do not misrepresent the world, they do not represent it at all.

Recognizing hallucination as a structural property of generative architectures has significant implications for the future of AI design and governance. It suggests that the search for more robust training techniques, larger datasets, or improved benchmarks cannot resolve the underlying problem. What is needed instead is an architectural reorientation, one that acknowledges the limits of generative coherence and incorporates external modules capable of grounding, constraint, and abstention. Hybrid systems, such as those enabled by Licensing Oracles, offer a path forward by restoring the distinction between linguistic fluency and epistemic responsibility.

More broadly, this analysis challenges the prevailing narrative in AI research that scaling and fine-tuning alone can yield models with reliable epistemic capabilities. As AI systems become more deeply embedded in scientific, legal, medical, and civic contexts, the distinction between coherence and truth becomes increasingly consequential. Designing systems that can participate responsibly in such contexts requires grappling with the ontological foundations of these models, not merely their behavioral outputs.

In reframing hallucination as an architectural and ontological issue rather than a behavioral or incentive-based one, we hope to reorient the discourse toward a deeper understanding of what generative models are and what they are not. Only by recognizing the limits imposed by their architecture can we build systems that are not merely fluent but trustworthy, systems that acknowledge the difference between language and world, and that respect it.